# BANGLABOOK: A Large-scale Bangla Dataset for Sentiment Analysis from Book Reviews

**Mohsinul Kabir**[*], **Obayed Bin Mahfuz**[*], **Syed Rifat Raiyan**[*],
**Hasan Mahmud, Md Kamrul Hasan**

Systems and Software Lab (SSL)
Department of Computer Science and Engineering
Islamic University of Technology, Dhaka, Bangladesh
{mohsinulkabir, siam, rifatraiyan, hasan, hasank}@iut-dhaka.edu

## Abstract

The analysis of consumer sentiment, as expressed through reviews, can provide a wealth of insight regarding the quality of a product. While the study of sentiment analysis has been widely explored in many popular languages, relatively less attention has been given to the Bangla language, mostly due to a lack of relevant data and cross-domain adaptability. To address this limitation, we present BANGLABOOK, a large-scale dataset of Bangla book reviews consisting of 158,065 samples classified into three broad categories: positive, negative, and neutral. We provide a detailed statistical analysis of the dataset and employ a range of machine learning models to establish baselines including SVM, LSTM, and Bangla-BERT. Our findings demonstrate a substantial performance advantage of pretrained models over models that rely on manually crafted features, emphasizing the necessity for additional training resources in this domain. Additionally, we conduct an in-depth error analysis by examining sentiment unigrams, which may provide insight into common classification errors in under-resourced languages like Bangla. Our codes and data are publicly available at https://github.com/mohsinulkabir14/BanglaBook.

## 1 Introduction

The resources publicly available for scholarly investigation in the realm of Sentiment Analysis (SA) for the Bangla language are scarce and limited in quantity (Khatun and Rabeya, 2022; Sazzed, 2021; Rahman et al., 2019) despite its literary gravitas as the 6$^{th}$ most spoken language[1] in the world with approximately 200 million speakers. In the existing literature on Bangla Text SA, as shown in Table 5, the largest dataset consists of 20,468 samples (Islam et al., 2022) while the smallest has a mere 1,050 samples (Tabassum and Khan, 2019). Besides these, Islam et al. (2020) created a dataset consisting of 17,852 samples and Islam et al. (2021) utilized a dataset of 15,728 samples. All other datasets apart from these either have <15,000 samples or are publicly unavailable. Another limitation of the existing research works in Bangla Text SA is the deficiency of datasets having product-specific review samples. Most of the available Bangla SA datasets are focused on user-generated textual content from cyberspace. The insights derived from these may not accurately represent sentiment in the context of product reviews, thus hindering their usefulness for businesses. The tonal and linguistic analysis of reviews from product-specific datasets can aid businesses to gain valuable insights into customer attitudes, preferences, and experiences which can then be leveraged to improve products and services, design targeted marketing campaigns, and make more informed business decisions. In this paper, we introduce a large-scale dataset, BANGLABOOK, consisting of 158,065 samples of book reviews collected from online bookshops written in the Bangla language. This is the largest dataset for Bangla sentiment analysis to the best of our knowledge. We perform an analysis of the dataset's statistical characteristics, employ various ML techniques to establish a performance benchmark for validating the dataset, and also conduct a thorough evaluation of the classification errors.

## 2 Dataset Construction

In order to create this dataset, we collect a total of 204,659 book reviews from two online bookshops (Rokomari[2] and Wafilife[3]) using a web scraper developed with several Python libraries, including BeautifulSoup, Selenium, Pandas, Openpyxl,

---

[*] These authors contributed equally to this work. Author names are in alphabetic order.
[1] https://en.wikipedia.org/wiki/List_of_languages_by_total_number_of_speakers
[2] https://www.rokomari.com/
[3] https://www.wafilife.com/

| Source | Language | Annotated | Unannotated | Total |
|---|---|---|---|---|
| Rokomari | Bangla | 84,672 | 24,806 | 109,478 |
| | Bangla† + English | 56,485 | 15,408 | 71,893 |
| | Bangla + Bangla† + English | 13,144 | 4,114 | 17,258 |
| | Total | 154,301 | 44,328 | 198,629 |
| Wafilife | Bangla | 4,699 | 59 | 4,758 |
| | Bangla† + English | 370 | 3 | 373 |
| | Bangla + Bangla† + English | 896 | 3 | 899 |
| | Total | 5,965 | 65 | 6,030 |
| | Bangla | 89,371 | 24,865 | 114,237 |
| | Bangla† + English | 56,855 | 15,411 | 72,266 |
| | Bangla + Bangla† + English | 14,040 | 4,117 | 18,157 |
| | Total | 160,266 | 44,393 | 204,659 |
| | Untranslated Data (Removed) | 2,201 | | |
| | Final Dataset Size | **158,065** | | |

Table 1: Summary statistics of our dataset. Bangla† denotes Romanized Bangla text.

and `Webdriver`, to collect and process the raw data.

For the data collection and preparation process of the BANGLABOOK dataset, we first compile a list of URLs for authors from online bookstores. From there, we procure URLs for the books. We meticulously scrape information such as book titles, author names, book categories, review texts, reviewer names, review dates, and ratings by utilizing these book URLs.

| | | |
|---|---|---|
| **General Properties** | # of Reviews | 158,065 |
| | # of Books | 30,253 |
| | # of Reviewers | 44,616 |
| | # of Categories | 1,573 |
| | Total Review Words | 44,429,201 |
| **Single Review** | max. Word length | 765 |
| | min. Word length | 1 |
| | avg. Word length | 281.08 |

Table 2: General overview of BANGLABOOK.

## 2.1 Labeling & Translation

If a review does not have a rating, we deem it unannotated. Reviews with a rating of 1 or 2 are classified as negative, a rating of 3 is considered neutral, and a rating of 4 or 5 is classified as positive. Two manual experiments are carried out to validate the use of ratings as a measure of sentiment in product reviews. In the first experiment, around 10% of the reviews are randomly selected and annotated manually. The annotated labels are cross-checked with the original labels, resulting in a 96.7% accuracy in the corresponding labels. In addition, we consult the work of Wang et al. (2020) that explored the issue of incongruous sentiment expressions with regard to ratings. Specifically, the study scrutinized two categories of reviews: *high ratings lacking a positive sentiment*, and *low ratings lacking a negative sentiment*. We perform an analysis to identify such inconsistencies within our dataset and discovered that only a minuscule 3.41% of the samples exhibited this pattern. This figure is relatively insignificant when considering the substantially large scale of our dataset.

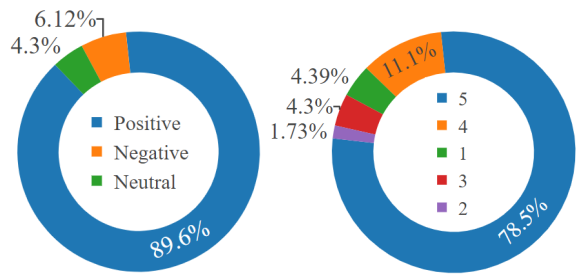

(a) Sentiment Distribution  (b) Rating Distribution

Figure 1: Class Distribution of BANGLABOOK.

After discarding the unannotated reviews, we curate a final dataset of 158,065 annotated reviews. Of these, 89,371 are written entirely in Bangla. The remaining 68,694 reviews were written in Romanized Bangla, English, or a mix of languages. They are translated into Bangla with Google Translator and a custom Python program using the `googletrans` library. The translations are subsequently subjected to manual review and scrutiny to confirm their accuracy. The majority of inaccurate translations primarily comprise spelling errors and instances where English words remain untranslated within samples containing a combination of Bangla and English text. The meticulous evaluation process of untranslated samples involves a thorough assessment by post-graduate native Bangla speakers, who critically compare the translated text against the original untranslated text to ascertain the correctness of the translation.

## 3 Statistical Analysis

Tables 1 and 2 provide an overview of the statistical properties of the BANGLABOOK dataset. The sentiment doughnut chart in Figure-1a illustrates the proportion of positive, neutral, and negative reviews, while the rating doughnut chart in Figure-1b displays the percentage of reviews that corre-spond to each rating on a scale of 1 to 5.

| Method | Negative | Neutral | Positive | Weighted Avg. |
|---|---|---|---|---|
| Random Forest(`word 2-gram + word 3-gram`) | 0.56 | 0.34 | 0.96 | 0.9106 |
| SVM(`word 2-gram + word 3-gram`) | 0.40 | 0.15 | **1.00** | 0.9053 |
| Random Forest(`word 1-gram`) | 0.48 | 0.35 | 0.96 | 0.9043 |
| Logistic Regression(`char 2-gram + char 3-gram`) | 0.55 | 0.13 | 0.96 | 0.8978 |
| Bangla-BERT(`base-uncased`) | 0.60 | 0.22 | 0.96 | 0.9064 |
| Logistic Regression(`word 2-gram + word 3-gram`) | 0.53 | 0.13 | 0.96 | 0.8964 |
| Bangla-BERT(`large`) | **0.72** | **0.40** | 0.97 | **0.9331** |
| XGBoost(`char 2-gram + char 3-gram`) | 0.31 | 0.02 | 0.95 | 0.8723 |
| Multinomial NB(`word 2-gram + word 3-gram`) | 0.23 | 0.03 | 0.95 | 0.8663 |
| LSTM(`GloVe`) | 0.11 | 0.00 | 0.10 | 0.0991 |
| XGBoost(`word 2-gram + word 3-gram`) | 0.23 | 0.01 | 0.95 | 0.8651 |
| Multinomial NB(`BoW`) | 0.18 | 0.05 | 0.94 | 0.8564 |
| SVM(`word 1-gram`) | 0.08 | 0.04 | 0.94 | 0.8519 |

Table 3: Catergory-wise Binary Task F1-score and Weighted Average F1-score of each method on BANGLABOOK.

Upon analyzing the sentiment chart, it appears that the majority of the reviews (124,084 + 17,503 = 141,587 samples) are positive, with a significant portion also being negative (2,728 + 6,946 = 9,674 samples). A relatively small fraction of the reviews are neutral (6,804 samples). This suggests that overall, the books have been well received by the readers, with the majority expressing favorable opinions. The distribution of the dataset is representative of real-world scenarios and it tessellates well with previous content analysis works on book reviews (Lin et al., 2005; Sorensen and Rasmussen, 2004). In Figure-2, we can visualize an illustration of the sentiment distribution among the 5 most frequently reviewed categories of books. We can gain some salient insights from the popularity of these genres. Contemporary novels are bestsellers as they reflect current events, social issues, and trends, making them relatable and thought-provoking for the readers while self-help and religious books provide guidance, inspiration, and a sense of purpose, catering to individuals' quest for personal growth and spiritual fulfillment.

## 4 Developing Benchmark for BANGLABOOK

A series of baseline models with combinations of different lexical and semantic features are chosen to evaluate the BANGLABOOK dataset. An overview of the models, evaluation metrics, results, and analysis of the experimental results are provided in this section.

### 4.1 Baseline Models & Features

For the lexical features, we extract bag-of-words (BoW), char $n$-grams (1-3), and word $n$-grams (1-3) from the reviews as these representations have performed well in different classification tasks (Islam et al., 2022). After extracting the features, they are vectorized using TF-IDF and count vec-torizer and trained on a series of ML models such as Random Forest (Breiman, 2001), XG-Boost (Chen and Guestrin, 2016), linear SVM (Cortes and Vapnik, 1995), Logistic Regression (le Cessie and van Houwelingen, 1992) and Multinomial Naive Bayes (John and Langley, 1995). We choose LSTM (Hochreiter and Schmidhuber, 1997) with GloVe (Pennington et al., 2014) embedding for its ability to understand context along with recent dependency. We also fine-tuned two available transformer-based models in Bangla: Bangla-BERT(`base-uncased`) (110M parameters) (Sarker, 2020) and Bangla-BERT(`large`) (pre-trained on 2.18B tokens) (Bhattacharjee et al., 2022), due to the recent success of BERT (Devlin et al., 2019) in various downstream NLP tasks. We select F1-score and weighted average F1-score to evaluate the models because the dataset has an un-even class distribution. F1-score is the harmonic mean of precision and recall and it helps balance the metric across the imbalanced positive/negative samples (Sokolova et al., 2006). All our experiments are done using scikit-learn, pytorch, and transformers (Vaswani et al., 2017) and run on Google Colaboratory. The training, testing, and validation split of the entire dataset was 70-20-10 with previously unseen samples in the test and validation set.

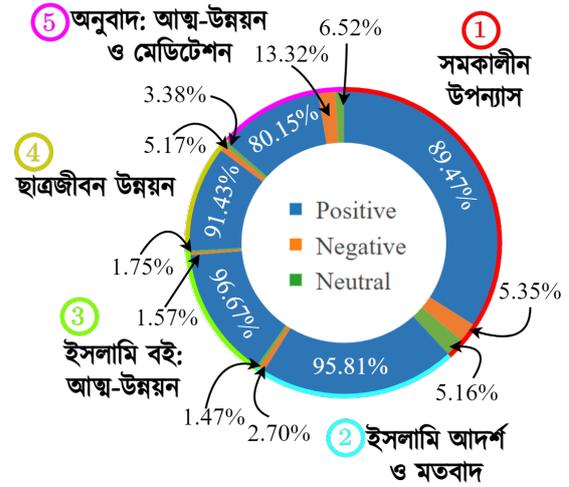

Figure 2: Sentiment Distribution of top 5 most popular genres. In clockwise order, সমকালীন উপন্যাস (Contemporary Novel), ইসলামি আদর্শ ও মতবাদ (Islamic Ideals and Doctrines), ইসলামি বই: আত্ম উন্নয়ন (Islamic Books: Self-Development), ছাত্রজীবন উন্নয়ন (Student Life Development), অনুবাদ: আত্ম-উন্নয়ন ও মেডিটেশন (Translated Books: Self-Development and Meditation).

| Class | 10 Most Frequent Words |
|---|---|
| Positive | ভালো (good), সুন্দর (nice), অসাধারণ (extraordinary), দুর্দান্ত (splendid), সেরা (best), সহজ (facile), চমৎকার (beautiful), আশা (hope), ধন্যবাদ (gratefulness), আলহামদুলিল্লাহ (gratitude) |
| Neutral | ভালো (good), সুন্দর (nice), খারাপ (bad), মোটামুটি (average), ভুল (fault), আশা (hope), সহজ (facile), অসাধারণ (extraordinary), কম (low), দুর্দান্ত (splendid) |
| Negative | ভালো (good), বাজে (trash), খারাপ (bad), ভুল (fault), সুন্দর (nice), আশা (hope), নষ্ট (waste), ফালতু (useless), হতাশা (disappointment), কম (low) |

Table 4: Most frequent word unigrams conveying the strongest sentiments of each class with English translation. The colors respectively denote Positive, Neutral and Negative sentiments.

Figure 3: Confusion Matrix for Bangla-BERT

### 4.2 Results & Findings

Table 3 summarizes the experimental results for BANGLABOOK. Results show that Bangla-BERT(large) outperforms all other models by a clear margin. Also, the combination of word/char 2-gram and word/char 3-gram perform exceptionally well with respective classifiers. Our hypothesis is that these two features result in a large number of unique word and character combinations, aiding the models' ability to generalize effectively across categories. Islam et al. (2022, 2021); Taher et al. (2018) concur on the same verdict by implying that the task predominantly relies on word units, with minimal dependence on subword level information and the nature of the Bangla language itself. Furthermore, Majumder et al. (2002) outlined the suitability of $n$-gram approaches in generating language profiles for Indo-European languages in their work. The bag-of-words (BoW) feature is inept at classifying the corresponding categories because of its inability to capture critical contextual information and nuance (Zheng and Casari, 2018). Although word 1-gram does not outperform word 2-gram and word 3-gram, it does predict the 'Neutral' class well. Both the pre-trained Bangla-BERT models perform fairly consistently across all categories on the BANGLABOOK dataset, demonstrating the usefulness of contextual understanding and transfer learning in classification tasks even in low-resource languages like Bangla. The LSTM model with GloVe embedding recognizes the 'Negative' and 'Positive' classes very marginally and fails completely to identify the 'Neutral' category. It is also notable that, SVM with bigram and trigram achieves perfect scores in the 'Positive' class.

To summarize, the utilization of pre-trained models (i.e. Bangla-BERT) that undergo training on extensive corpora, leading to exposure to extensive general language knowledge, has significantly contributed to their superior classification performance compared to other models and word embeddings. Additionally, models trained on hand-crafted features also perform significantly well. It should be noted that Bangla pre-trained models are currently undergoing development, and further training on expansive corpora has the potential to enhance their ability to generalize and achieve even more impressive results.

### 4.3 Error Analysis

In the 'Positive' class, all the models produce excellent classification results. While some models perform reasonably well on the 'Negative' class, nearly all of the models perform poorly on the 'Neutral' class. The class imbalance of the dataset, as shown in Figure 1, is one obvious cause of this fluctuation in results. The confusion matrix for

Bangla-BERT on our dataset, presented in Figure-3, reveals that most of the 'Negative' and 'Neutral' samples are misclassified as 'Positive' samples by our classifiers. To further analyze the misclassifications, we examine the W1 (word unigrams) of these three classes. We find 124,796 unique W1 for the 'Positive' class, 20,714 unique W1 for the 'Negative' class, and 19,096 unique W1 for the 'Neutral' class. 77.57% of the W1 from the 'Neutral' class and 79.83% of the W1 from the 'Negative' class are found in the 'Positive' class. Table 4 depicts the most frequent W1 conveying the strongest sentiments for each class. With only one distinct 'Neutral' W1 and even the 'Negative' class having multiple positive W1, the dominance of 'Positive' sentiment W1 over the other two classes is evident. This may have contributed to the lack of distinctive words in the 'Negative' and 'Neutral' classes, which inevitably prevented the feature-based models from generalizing.

## 5 Morphology and Negation Patterns of Bangla

Understanding the morphology and negation patterns of a language holds paramount importance in the realm of sentiment analysis because negation can alter the meaning of words and phrases, thereby affecting the overall sentiment conveyed by a text. We provide a concise yet insightful recapitulation of the topic in the case of Bangla accompanied by review samples from our dataset BANGLABOOK as the respective examples. From the linguistic typological standpoint, Bangla is categorized as a *subject-object-verb* (SOV) language because the subject, object, and verb generally adhere to said order in its sentential structure (Ramchand, 2004). The most common juxtaposition of polarity from positive to negative is the use of *ni* (নি) as a tensed negative. For example,

আমি তাঁর অনুরাগী হওয়ায় আমি এই বইটি কেনা থেকে নিজেকে প্রতিরোধ করতে পারিনি !!!!
**Translation:** As I am a fan of his I couldn't resist myself from buying this book!!!!

Another negational feature is expressed by placing *na* (না) prior to the non-finite verb and after the finite verb in a sentence (although there are some exceptions). For example,

অবশ্য হুমায়ুন আহমেদ লিখেছেন এই বইটার উপরে তিনি নিজেও সন্তুষ্ট না ।
**Translation:** Of course, Humayun Ahmed wrote that he himself is not satisfied with this book.

The Bangla language consists of no negative adverbs or pronouns (Thompson, 2006). This is why the negative element responsible for the reversal of polarity transcends from the word-level to the sentence-level rendering the occurrences of almost all negations in Bangla manifest on the syntactic level (Thompson, 2006).

In the cases of double negatives, we see the involvement of lexical negation, a morphological feature that works with negative affixes (prefixes and suffixes) attached to a root word. The prefixes in Bangla have two different phonetic variations or allophones depending on whether the prefix precedes a vowel or a consonant. The same is true for prefixes that imbue a negative connotation to a root word, e.g. *o* (অ) and *on* (অন্). For example,

কিন্তু এই বইটি এই অপূর্ণতা ঢেকে ফেলেছে।
**Translation:** But this book has covered up this incompleteness .

ওমর খৈয়ামের ভাষায় কিছু বই অনন্ত যৌবনের বই, যাদের কোন ক্ষয় নেই।
**Translation:** In the words of Omar Khayyam, some books are books of never-ending youth, which have no decay.

Another negative prefix that precedes a root word to invert its polarity is *nir* (নির্). For example,

লেখকের নিরলস শ্রম লেখায় ফুটে উঠেছে।
**Translation:** The relentless effort of the author is reflected in the writing.

On the contrary, the suffix *hin* (হীন) succeeds a root word to convert it to the corresponding negative form. For example,

এরকম ভিত্তিহীন কাল্পনিক গল্প শিশুদের না পড়াই ভালো।
**Translation:** It is better for children not to read such baseless fictional stories.

The expression of negative sentiment is, therefore, very nuanced in the Bangla language as every occurrence of negative is intertwined with features like the tense, hierarchy of syntax, verb status, case-specific issues, and sequential arrangement of words (Thompson, 2006).

## 6 Conclusion

This paper introduces BANGLABOOK, the largest Bangla book review dataset with 158,065 samples, each labeled with 1 of 3 user sentiments. We provide extensive statistical analysis and strong baselines facilitating the utility of the dataset. Given its massive size and fine-grained sentiment distribution, BANGLABOOK has the potential to alleviate the resource scarcity in Bangla language research.

# 7 Limitations

Many of the reviews that were gathered for constructing BANGLABOOK are discarded because they lack a corresponding rating. A manual annotation process would have yielded a much larger dataset, which was not feasible due to resource constraints. Moreover, one of the challenges for validating the dataset is the lack of statistical models and word-embeddings pre-trained on the Bangla language. Some pre-trained Bangla-BERT models, yet to be trained on extensive corpora, have only recently been proposed. Improving transformer-based models for Bangla can enhance sub-word level contextual understanding which will consequently help in more accurate identification of the sentiments in BANGLABOOK (Islam et al., 2022).

# References

Mst Tuhin Akter, Manoara Begum, and Rashed Mustafa. 2021. Bengali sentiment analysis of e-commerce product reviews using k-nearest neighbors. In *2021 International Conference on Information and Communication Technology for Sustainable Development (ICICT4SD)*, pages 40–44. IEEE.

Sanjida Akter and Muhammad Tareq Aziz. 2016. Sentiment analysis on facebook group using lexicon based approach. In *2016 3rd International Conference on Electrical Engineering and Information Communication Technology (ICEEICT)*, pages 1–4. IEEE.

Shad Al Kaiser, Sudipta Mandal, Ashraful Kalam Abid, Ekhfa Hossain, Ferdous Bin Ali, and Intisar Tahmid Naheen. 2021. Social media opinion mining based on bangla public post of facebook. In *2021 24th International Conference on Computer and Information Technology (ICCIT)*, pages 1–6. IEEE.

Abhik Bhattacharjee, Tahmid Hasan, Wasi Ahmad, Kazi Samin Mubasshir, Md Saiful Islam, Anindya Iqbal, M. Sohel Rahman, and Rifat Shahriyar. 2022. BanglaBERT: Language model pretraining and benchmarks for low-resource language understanding evaluation in Bangla. In *Findings of the Association for Computational Linguistics: NAACL 2022*, pages 1318–1327, Seattle, United States. Association for Computational Linguistics.

L. Breiman. 2001. Random forests. *Machine Learning*, 45:5–32.

Tianqi Chen and Carlos Guestrin. 2016. Xgboost: A scalable tree boosting system. *Proceedings of the 22nd ACM SIGKDD International Conference on Knowledge Discovery and Data Mining*.

Shaika Chowdhury and Wasifa Chowdhury. 2014. Performing sentiment analysis in bangla microblog posts. In *2014 International Conference on Informatics, Electronics & Vision (ICIEV)*, pages 1–6. IEEE.

Corinna Cortes and Vladimir Naumovich Vapnik. 1995. Support-vector networks. *Machine Learning*, 20:273–297.

Jacob Devlin, Ming-Wei Chang, Kenton Lee, and Kristina Toutanova. 2019. Bert: Pre-training of deep bidirectional transformers for language understanding. *ArXiv*, abs/1810.04805.

Rajib Chandra Dey and Orvila Sarker. 2019. Sentiment analysis on bengali text using lexicon based approach. In *2019 22nd International Conference on Computer and Information Technology (ICCIT)*, pages 1–5. IEEE.

Asif Hassan, Mohammad Rashedul Amin, Abul Kalam Al Azad, and Nabeel Mohammed. 2016. Sentiment analysis on bangla and romanized bangla text using deep recurrent models. In *2016 International Workshop on Computational Intelligence (IWCI)*, pages 51–56. IEEE.

Sepp Hochreiter and Jürgen Schmidhuber. 1997. Long short-term memory. *Neural Computation*, 9:1735–1780.

MD Iqbal, Avishek Das, Omar Sharif, Mohammed Moshiul Hoque, and Iqbal H Sarker. 2022. Bemoc: A corpus for identifying emotion in bengali texts. *SN Computer Science*, 3(2):1–17.

Khondoker Ittehadul Islam, Md Saiful Islam, and Md Ruhul Amin. 2020. Sentiment analysis in bengali via transfer learning using multi-lingual bert. In *2020 23rd International Conference on Computer and Information Technology (ICCIT)*, pages 1–5. IEEE.

Khondoker Ittehadul Islam, Sudipta Kar, Md Saiful Islam, and Mohammad Ruhul Amin. 2021. Sentnob: A dataset for analysing sentiment on noisy bangla texts. In *Findings of the Association for Computational Linguistics: EMNLP 2021*, pages 3265–3271.

Khondoker Ittehadul Islam, Tanvir Yuvraz, Md Saiful Islam, and Enamul Hassan. 2022. Emonoba: A dataset for analyzing fine-grained emotions on noisy bangla texts. In *Proceedings of the 2nd Conference of the Asia-Pacific Chapter of the Association for Computational Linguistics and the 12th International Joint Conference on Natural Language Processing*, pages 128–134.

George H. John and Pat Langley. 1995. Estimating continuous distributions in bayesian classifiers. In *Conference on Uncertainty in Artificial Intelligence*.

Mst Eshita Khatun and Tapasy Rabeya. 2022. A machine learning approach for sentiment analysis of book reviews in bangla language. In *2022 6th International Conference on Trends in Electronics and Informatics (ICOEI)*, pages 1178–1182. IEEE.

Saskia le Cessie and J. C. van Houwelingen. 1992. Ridge estimators in logistic regression. *Applied statistics*, 41:191–201.

Tom MY Lin, Pin Luarn, and Yun Kuei Huang. 2005. Effect of internet book reviews on purchase intention: A focus group study. *The Journal of Academic Librarianship*, 31(5):461–468.

Shamsul Arafin Mahtab, Nazmul Islam, and Md Mahfuzur Rahaman. 2018. Sentiment analysis on bangladesh cricket with support vector machine. In *2018 international conference on Bangla speech and language processing (ICBSLP)*, pages 1–4. IEEE.

P Majumder, M Mitra, and BB Chaudhuri. 2002. Ngram: a language independent approach to ir and nlp. In *International conference on universal knowledge and language*, volume 2.

Muhammad Mahmudun Nabi, Md Tanzir Altaf, and Sabir Ismail. 2016. Detecting sentiment from bangla text using machine learning technique and feature analysis. *International Journal of Computer Applications*, 153(11):28–34.

Jeffrey Pennington, Richard Socher, and Christopher D. Manning. 2014. Glove: Global vectors for word representation. In *Conference on Empirical Methods in Natural Language Processing*.

Md Rahib, Rumman Hussain Khan, Amzad Hussain Tamim, Mohammad Zawad Tahmeed, and Mohammad Jaber Hossain. 2022. Emotion detection based on bangladeshi peoples social media response on covid-19. *SN Computer Science*, 3(2):1–6.

Fuad Rahman, Habibur Khan, Zakir Hossain, Mahfuza Begum, Sadia Mahanaz, Ashraful Islam, and Aminul Islam. 2019. An annotated bangla sentiment analysis corpus. In *2019 International Conference on Bangla Speech and Language Processing (ICBSLP)*, pages 1–5. IEEE.

Md Atikur Rahman and Emon Kumar Dey. 2018. Datasets for aspect-based sentiment analysis in bangla and its baseline evaluation. *Data*, 3(2):15.

Gillian Catriona Ramchand. 2004. Two types of negation in bengali. *Clause structure in South Asian languages*, pages 39–66.

Sagor Sarker. 2020. Banglabert: Bengali mask language model for bengali language understanding.

Salim Sazzed. 2020a. Cross-lingual sentiment classification in low-resource bengali language. In *Proceedings of the sixth workshop on noisy user-generated text (W-NUT 2020)*, pages 50–60.

Salim Sazzed. 2020b. Development of sentiment lexicon in bengali utilizing corpus and cross-lingual resources. In *2020 IEEE 21st International conference on information reuse and integration for data science (IRI)*, pages 237–244. IEEE.

Salim Sazzed. 2021. Bengsentilex and bengswearlex: creating lexicons for sentiment analysis and profanity detection in low-resource bengali language. *PeerJ Computer Science*, 7:e681.

Marina Sokolova, Nathalie Japkowicz, and S. Szpakowicz. 2006. Beyond accuracy, f-score and roc: A family of discriminant measures for performance evaluation. In *Australian Conference on Artificial Intelligence*.

Alan T Sorensen and Scott J Rasmussen. 2004. Is any publicity good publicity? a note on the impact of book reviews. *NBER Working paper, Stanford University*.

Nusrath Tabassum and Muhammad Ibrahim Khan. 2019. Design an empirical framework for sentiment analysis from bangla text using machine learning. In *2019 International Conference on Electrical, Computer and Communication Engineering (ECCE)*, pages 1–5. IEEE.

SM Abu Taher, Kazi Afsana Akhter, and KM Azharul Hasan. 2018. N-gram based sentiment mining for bangla text using support vector machine. In *2018 international conference on Bangla speech and language processing (ICBSLP)*, pages 1–5. IEEE.

Hanne-Ruth Thompson. 2006. Negation patterns in bengali. *Bulletin of the School of Oriental and African Studies*, 69(2):243–265.

Rashedul Amin Tuhin, Bechitra Kumar Paul, Faria Nawrine, Mahbuba Akter, and Amit Kumar Das. 2019. An automated system of sentiment analysis from bangla text using supervised learning techniques. In *2019 IEEE 4th International Conference on Computer and Communication Systems (ICCCS)*, pages 360–364. IEEE.

Ashish Vaswani, Noam M. Shazeer, Niki Parmar, Jakob Uszkoreit, Llion Jones, Aidan N. Gomez, Lukasz Kaiser, and Illia Polosukhin. 2017. Attention is all you need. *ArXiv*, abs/1706.03762.

Anning Wang, Qiang Zhang, Shuangyao Zhao, Xiaonong Lu, and Zhanglin Peng. 2020. A review-driven customer preference measurement model for product improvement: sentiment-based importanceperformance analysis. *Information Systems and e-Business Management*, 18:61–88.

Alice Zheng and Amanda Casari. 2018. *Feature engineering for machine learning: principles and techniques for data scientists*. " O'Reilly Media, Inc.".

# A  Appendix

| Dataset | Sentiment Classification | Sentiment Distribution | | Total # of Samples | Availability | Type of Content | Source(s) | Baseline Models |
|---|---|---|---|---|---|---|---|---|
| (Tabassum and Khan, 2019) | Positive<br>Negative | -<br>- | | 1,050 | Closed | Posts, comments | Facebook, Twitter | RF |
| (Chowdhury and Chowdhury, 2014) | Positive<br>Negative | -<br>- | | 1,300 | Closed | Posts, comments | Twitter | - |
| (Nabi et al., 2016) | Positive<br>Negative | -<br>- | | 1,500 | Closed | Posts, comments | Social Media | - |
| (Mahtab et al., 2018) | Praise<br>Criticism<br>Sadness | 513<br>604<br>484 | | 1,601 | Closed | Comments | Prothom Alo Online News Portal | SVM, DT, NB |
| (Akter and Aziz, 2016) | Positive<br>Negative<br>Neutral | -<br>-<br>- | | 3,600 | Closed | Posts, comments | Facebook | NB |
| (Rahman and Kumar Dey, 2018) | Positive<br>Negative<br>Neutral | -<br>-<br>- | | 4,700 | **Open** | Comments | Facebook pages: BBC Bangla, Prothom Alo | SVM, LR, KNN, DT, LSTM NB, CNN |
| (Dey and Sarker, 2019) | Positive<br>Negative | 2,600<br>2,600 | | 5,200 | Closed | Comments, reviews | Facebook, Twitter, YouTube, News Portals | DT, NB, SVM |
| (Khatun and Rabeya, 2022) | Positive<br>Negative | -<br>- | | 5,500 | Closed | Comments, reviews | Social Media | - |
| BEMOC (Iqbal et al., 2022) | Anger<br>Fear<br>Surprise<br>Sadness<br>Joy<br>Disgust | -<br>-<br>-<br>-<br>-<br>- | | 7,000 | **Open** | Posts, comments | Facebook, YouTube, Online blogs, Bangla story books, novels, newspapers, discourse | - |
| (Tuhin et al., 2019) | Happy<br>Tender<br>Excited<br>Sad<br>Angry<br>Scared | -<br>-<br>-<br>-<br>-<br>- | | 7,500 | Closed | - | - | NB, Tropical Method |
| (Akter et al., 2021) | Positive<br>Negative<br>Neutral | -<br>-<br>- | | 7,905 | Closed | Product reviews | Daraz | RF, LR, SVM, KNN, XGB |
| (Hassan et al., 2016) | Positive<br>Negative<br>Ambiguous | -<br>-<br>- | | 9,337 | Closed | Comments, reviews | Facebook, Twitter, YouTube, News Portals | LSTM |
| (Rahib et al., 2022) | Insightful<br>Curious<br>Gratitude | 3,800<br>3,549<br>3,232 | | 10,581 | Closed | Comments | Social Media | SVM, RF, CNN, LSTM |
| (Al Kaiser et al., 2021) | Positive | Wishful Thinking<br>Appreciation | 967<br>942 | 1,909 | 11,006 | Closed | Comments | Facebook | LR, DT, RF, MNB, KNN, Linear SVM, RBF SVM, XGB |
| | Negative | Gender-based hate<br>Religious hate<br>Political hate<br>Personal hate<br>Sarcasm | 525<br>731<br>572<br>1,995<br>1,414 | 5,237 | | | | | |
| | Neutral | - | 3,860 | 3,860 | | | | | |
| (Sazzed, 2020a) | Positive<br>Negative | 8,500<br>3,307 | | 11,807 | **Open** | Comments | YouTube | SVM, ET, RF, LR, VADER, TextBlob |
| (Sazzed, 2020b) | Positive<br>Negative | -<br>- | | 12,000 | Closed | Comments | YouTube | - |
| SENTNOB (Islam et al., 2021) | Positive<br>Negative<br>Neutral | 6,410<br>5,709<br>3,609 | | 15,728 | **Open** | Comments | Prothom Alo Online Newspaper, YouTube | RNN |
| (Islam et al., 2020) | Positive<br>Negative<br>Neutral | 4,769<br>8,351<br>4,732 | | 17,852 | **Open** | Comments | Prothom Alo Online Newspaper | CNN, LSTM, BERT, GRU, fastText |
| EMONOBA (Islam et al., 2022) | Love<br>Joy<br>Surprise<br>Anger<br>Sadness<br>Fear | 4,202<br>9,249<br>939<br>3,905<br>5,109<br>307 | | 20,468 | **Open** | Comments | YouTube, Facebook, Twitter, Prothom Alo | Bi-LSTM, fastText, Bangla-BERT-base |
| BANGLABOOK (ours) | Positive<br>Negative<br>Neutral | 141,587<br>9,674<br>6,804 | | **158,065** | **Open**[†] | Book reviews | Rokomari, Wafilife | RF, LSTM, LR, GRU MNB, SVM, XGB Bangla-BERT |

Table 5: Comparison of notable Bangla Sentiment Analysis datasets sorted in ascending order of size. The abbreviations respectively denote, Random Forest (RF), Support Vector Machine (SVM), Decision Tree (DT), Naive Bayes (NB), Logistic Regression (LR), K-Nearest Neighbors (KNN), Long Short-term Memory (LSTM), Convolutional Neural Network (CNN), Extreme Gradient Boost (XGB), Multinomial Naive Bayes (MNB), Radial Basis Function (RBF), Extreme Random Tree (ET), Recurrent Neural Network (RNN), Bidirectional Encoder Representations from Transformers (BERT), Gated Recurrent Unit (GRU), Bi-LSTM (Bidirectional LSTM). All the publicly available datasets are hyperlinked. Open[†] denotes the redaction of the link for anonymity.